\documentclass[10pt,twocolumn,letterpaper]{article}

\usepackage{iccv}              
\usepackage{algorithm}
\usepackage{algorithmic}
\usepackage{amsmath}

\usepackage{amssymb}
\usepackage{bbding}
\usepackage{pifont}
\usepackage{xcolor}
\usepackage{tikz}
\definecolor{cjcolor}{RGB}{0,0,255}

\definecolor{iccvblue}{rgb}{0.21,0.49,0.74}
\usepackage[pagebackref,breaklinks,colorlinks,allcolors=iccvblue]{hyperref}

\newcommand\blfootnote[1]{%
\begingroup
\renewcommand\thefootnote{}\footnote{#1}%
\addtocounter{footnote}{-1}%
\endgroup
}

\def\paperID{10581} 
\def\confName{ICCV}
\def\confYear{2025}

\title{DocTron-Formula: Generalized Formula Recognition in Complex and Structured Scenarios}

\author{Yufeng Zhong \quad Zhixiong Zeng$^{\dagger}$ \quad Lei Chen \quad Longrong Yang \quad Liming Zheng \\ Jing Huang \quad Siqi Yang \quad Lin Ma$^{\ddagger}$ \\ \\
Meituan}

\begin{document}

\maketitle

\blfootnote{$\dagger$ Project leader. $\ddagger$ Corresponding author.}

\begin{abstract}
Optical Character Recognition (OCR) for mathematical formula is essential for the intelligent analysis of scientific literature. However, both task-specific and general vision-language models often struggle to handle the structural diversity, complexity, and real-world variability inherent in mathematical content. In this work, we present DocTron-Formula, a unified framework built upon general vision-language models, thereby eliminating the need for specialized architectures. Furthermore, we introduce CSFormula, a large-scale and challenging dataset that encompasses multidisciplinary and structurally complex formulas at the line, paragraph, and page levels. Through straightforward supervised fine-tuning, our approach achieves state-of-the-art performance across a variety of styles, scientific domains, and complex layouts. Experimental results demonstrate that our method not only surpasses specialized models in terms of accuracy and robustness, but also establishes a new paradigm for the automated understanding of complex scientific documents. Available at: \url{https://github.com/DocTron-hub/DocTron-Formula}
\end{abstract}
    
\section{Introduction}
\label{sec:intro}
Optical Character Recognition (OCR) for mathematical formulas plays a fundamental role in the digitization and intelligent processing of scientific literature. It enables the automatic retrieval, editing, and analysis of complex academic content, thereby providing robust support for academic research and knowledge management. 

In recent years, a variety of specialized models have been proposed for formula recognition~\cite{wang2024unimernet, xia2025latexnet, liu2025pp, wang2025enhancing, guan2024posformer, yang2023read}, achieving notable progress in accuracy and efficiency. Despite significant progress in this field, existing methods still face notable challenges in several critical aspects, making it difficult to fully address the requirements of real-world scientific literature processing:

\noindent \textbf{(1) Reliance on Task-specific Models.} Most existing approaches for formula recognition are based on task-specific models that require carefully designed architectures. While recent advances in general vision-language models (VLMs), such as GPT-4o~\cite{openai2024gpt4o}, have enabled formula recognition without the need for specialized model design, their performance still lags behind that of dedicated models. This highlights the need for a truly generalizable model that can achieve competitive performance with task-specific solutions, while avoiding the burden of customized architecture for each new scenario.

\begin{figure*}[!htbp]
\centering
\includegraphics[width=\textwidth]{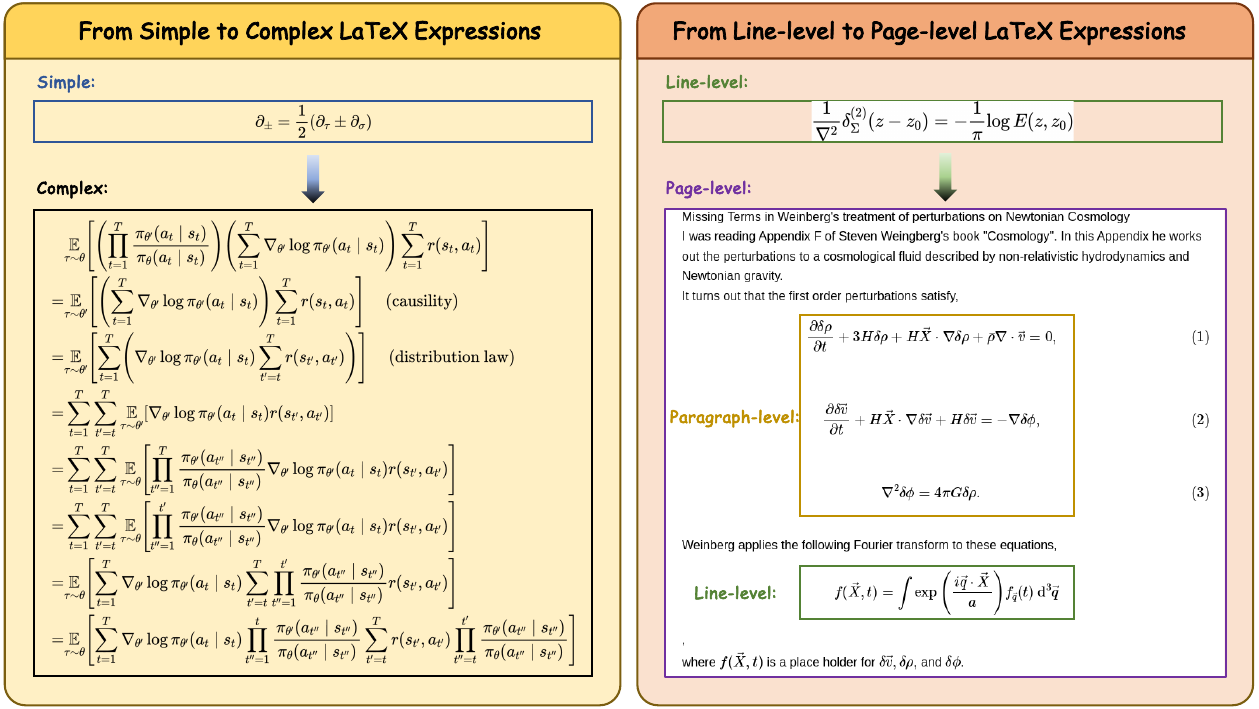}
\vspace{-6mm}
\caption{
Illustration of the main challenges in current LaTeX formula datasets: limited complexity and structural diversity. \textit{Left}: \textbf{Simple vs. Complex LaTeX Expressions.} Existing datasets primarily contain simple, single-structure formulas with few symbol types, which restricts model performance on advanced scientific content. In contrast, the proposed CSFormula dataset includes complex expressions with multi-level nesting, sophisticated subscripts and superscripts, and abundant scientific symbols, supporting more accurate recognition of diverse formulas. \textit{Right}: \textbf{Line-level vs. Page-level LaTeX Expressions.} Most datasets focus only on line-level formulas, overlooking the complex paragraph- and page-level layouts common in real documents. The proposed CSFormula dataset incorporates these higher-level structures, enabling models to better capture complex layouts and improving semantic reconstruction and document understanding.
}
\vspace{-2mm}
\label{fig:motivation}
\end{figure*}

\noindent \textbf{(2) Insufficient Difficulty and Complexity.} As illustrated at the \textit{left} of Fig.~\ref{fig:motivation}, mainstream methods and public datasets predominantly focus on structurally simple formulas with limited symbol diversity. However, they pay insufficient attention to authentic scientific formulas that are multidisciplinary, highly complex, and structurally diverse. Consequently, existing models encounter significant bottlenecks when processing deeply nested structures, intricate superscripts and subscripts, special operators, and cross-disciplinary symbol systems. This limitation hinders their ability to effectively handle the rich structures and diverse expressions present in advanced formulas across fields such as physics, chemistry, and biology.

\noindent \textbf{(3) Limited Structural Adaptability and Generalization.} As illustrated at the \textit{right} of Fig.~\ref{fig:motivation}, most existing methods focus on single-line formulas, often relying on tools such as PDF-Extract-Kit\footnote{\url{https://github.com/opendatalab/PDF-Extract-Kit}} to detect and extract formula regions from PDFs or webpage screenshots. However, they pay insufficient attention to multi-line formulas and the complex, page-level layouts that are prevalent in real-world documents. 
As a result, these specialized approaches often lack robustness when applied to real-world documents, leading to performance degradation and limiting their effectiveness in accurately restoring formula semantics and supporting automatic document understanding.

To effectively address these challenges, we propose a comprehensive solution.
First, we introduce DocTron-Formula, a unified framework based on a general VLM, which eliminates the need for task-specific architectures. This large-model-driven approach enables efficient adaptation to diverse and complex formula recognition scenarios through simple fine-tuning.
Second, to overcome the limited difficulty and structural simplicity of existing datasets, and to improve structural adaptability and generalization, we construct a Complex and Structured Formula dataset (CSFormula). The CSFormula dataset covers multidisciplinary, highly complex, and structurally diverse formulas, including not only single-line formulas but also multi-line and page-level complex layouts, enabling models to learn rich structures and varied expressions present in advanced scientific fields and ensuring robust performance in real-world document scenarios.
Finally, our method achieves state-of-the-art (SOTA) performance across different styles, disciplines, and page structures, fully validating the superior accuracy and broad applicability of our approach.

Specifically, our method demonstrates significant innovation and advantages in the following aspects:

\noindent \textbf{(1) Construction of High-Difficulty and Multi-Structural Datasets.} We have independently designed and implemented efficient data acquisition and processing tools to systematically collect, clean, and organize a large number of complex formula samples from academic resources across multiple disciplines. The CSFormula dataset covers mathematics, physics, chemistry, and other fields, as well as multi-line and page-level complex layout structures, thereby more authentically reflecting the diversity and challenges of formulas found in the literature and providing a solid foundation for model training and evaluation.

\noindent \textbf{(2) General VLM-Driven Complex Formula Recognition.} Unlike traditional specialized models that rely on structural customization or domain-specific architectures, we directly leverage general large-scale multimodal pre-trained models such as Qwen2.5-VL~\cite{Qwen2.5-VL}. Through supervised fine-tuning (SFT) on the high-difficulty dataset for domain adaptation, experimental results show that large models, owing to their strong knowledge transfer and structural generalization capabilities, can achieve SOTA recognition performance in various complex scenarios with only simple fine-tuning, without the need for complex engineering or extensive manual rules. This demonstrates the inherent advantages and broad application prospects of general large models in complex formula recognition tasks.

\noindent \textbf{(3) Superior Performance.} Our method demonstrates extremely high recognition accuracy and stability across different styles (such as printed, handwritten, and screenshot), varying disciplinary difficulties, and diverse complex page structures. It achieves SOTA results on three public datasets, further validating the superiority and generalizability of our approach.

In summary, through the construction of high-quality datasets, general large-model-driven complex formula recognition, our approach systematically demonstrates for the first time the outstanding performance of general multimodal large models in the recognition of high-difficulty, structurally complex formulas. This not only provides a novel paradigm for the automated understanding of complex scientific literature, but also lays a solid foundation for further research and applications in related fields.

\section{Related Work}
\label{sec:related_work}

\subsection{Formula Recognition}
Formula recognition is a long-standing challenge with the first study dating to~\cite{anderson1967syntax}. 
Initially, some hand-crafted grammar rules are used to model the spatial structure of formulas, \textit{e.g.}, graph grammars~\cite{LavirotteP98}, relational grammars~\cite{maclean2013new}, and probabilistic grammars~\cite{awal2014global,alvaro2016integrated}.
Then, the CROHME competitions~\cite{mouchere2013icdar,le2019pattern,mouchere2014icfhr,mouchere2016icfhr2016,mahdavi2019icdar} play a pivotal role in advancing deep learning algorithms for formula recognition. 
Key works include a neural encoder-decoder model featuring coarse-to-fine attention mechanisms~\cite{deng2017image}, a tree-structured decoder~\cite{zhang2020tree}, and the Counting-Aware Network~\cite{li2022counting}.
More recently, the rapid evolution of Transformer~\cite{vaswani2017attention} has spurred researchers to explore formula OCR supported by carefully designed evaluation benchmarks such as DocGenome~\cite{xia2024docgenome} and MMSci~\cite{li2024mmsci}.
For instance, Donut~\cite{kim2022ocr} introduces an end-to-end model that converts document images into structured outputs. 
Nougat~\cite{blecher2023nougat} uses automatically generated image-to-markup samples to train a Transformer-based encoder-decoder model. 
Vary~\cite{wei2023vary} offers a fine-grained multimodal solution for document parsing. Additionally, Pix2tex~\cite{pix2tex2022} and Texify~\cite{texify2023} specifically address the unique challenges of mathematical expressions.

A trend in formula recognition is the growing complexity of formulas.
UniMERNet~\cite{wang2024unimernet} introduces the UniMER dataset, the first to target complex real-world Mathematical Expression Recognition (MER), and proposes UniMERNet as a universal MER framework. 
LaTeXNet~\cite{xia2025latexnet} focuses on LaTeX transcoding for visual tables and formulas, achieving efficient formula parsing via a specially designed model structure.
PP-FormulaNet~\cite{liu2025pp} strikes a balance between accuracy and efficiency in recognizing complex formulas. 
HD-Net~\cite{wang2025enhancing} presents a hierarchical detail-focused mechanism to deepen the understanding of multi-level expressions. 
PosFormer~\cite{guan2024posformer} enhances the parsing capability of handwritten mathematical expressions through structural relationship modeling. TenL~\cite{yang2023read} develops a row-aware semi-autoregressive Transformer to enable efficient parsing of multi-line handwritten formulas.
However, existing research still overlooks the interleaved structures and intricate relationships between text and formulas in real-world manuscripts, limiting their ability to fully meet the demands of scientific literature processing.

\subsection{Large Vision-language Models}
Large Language Models (LLMs) have showcased remarkable capabilities in following instructions and generalizing across tasks. 
However, a key limitation is that LLMs can only process textual information, whereas real-world applications demand models capable of handling visual data. 
To reduce this gap, Large Vision-Language Models (LVLMs) incorporate visual information by leveraging frozen visual encoders and trainable visual projectors to integrate visual data into LLMs.
Recent research has focused on enhancing LVLMs through two types of methods. 
The first optimizes training strategies, \textit{e.g.},~\cite{bai2023qwenvl, chen2023minigpt,bai2025qwen2}. 
The second, and more prevalent, approach centers on improving visual components: this includes expanding visual instruction-tuning datasets ~\cite{liu2023aligning, zhang2023llavar}, advancing image encoders ~\cite{chen2023internvl, bai2023qwenvl}, and refining the alignment between input and projection layers ~\cite{lin2023video, cha2023honeybee, alayrac2022flamingo, dai2023instructblip, ye2023mplug, zhao2023svit}. These efforts have significantly boosted visual understanding capabilities of LVLMs.
Besides, some studies have adapted large vision-language models to enhance document parsing abilities ~\cite{wang2024mineru, li2025monkeyocr, wang2025marten, yu2024texthawk2}.
Despite their notable success in general vision-language tasks, it is underexplored that how these models perform in formula recognition tasks, especially considering the high symbol density of complex formulas and their discipline-specific properties.

\section{Methodology}
\label{sec:method}

\begin{table*}[!htbp]
\centering
\small
\resizebox{0.6\textwidth}{!}{
\begin{tabular}{l|c|c|c}
\hline
\textbf{Level} & \textbf{Before Deduplication} & \textbf{Train} & \textbf{Test} \\
\hline
Line-level     & 1,884,532 & 741,016  & 1,000 \\
Paragraph-level &   412,789 & 135,575  & 1,000 \\
Page-level     &   412,789 & 131,876  & 1,000 \\
\hline
\end{tabular}
}
\vspace{-1mm}
\caption{Statistics of the dataset at line, paragraph, and page levels, showing the number of samples before deduplication as well as the split between the training and test sets after deduplication.}
\vspace{-2mm}
\label{tab:dedup}
\end{table*}

\subsection{Overview}
Formula recognition is an important research topic, providing strong support for academic research and knowledge management.
As shown in Figure~\ref{fig:motivation}, existing formula recognition methods face two main challenges: $(i)$ Formulas have limited complexity and poor structural diversity.
Formulas typically include complex expressions with multi-level nesting and abundant scientific symbols.
Existing public datasets fail to account for this, making them unable to handle the recognition of advanced formulas in real-world scenarios. 
Besides, formulas often interleave with text, and multiple formulas on a page are frequently interconnected, a scenario not covered by existing public datasets. 
This leads to issues such as loss of structural information and incorrect symbol associations in real-world applications. 
$(ii)$ Formula recognition methods rely on structural customization and extensive prior knowledge, which limits the generalization ability of these models across different formula recognition scenarios.

To this end, we propose a systematic solution.
First, to address the limited complexity and insufficient structural diversity of existing formula datasets, we have constructed a large-scale, high-difficulty dataset, CSFormula, which covers multiple disciplines, diverse structures, and complex layouts.
Second, moving beyond existing methods that rely on specific structural modeling, we adopt a general VLM-driven approach, DocTron-Formula, for complex formula recognition. This approach can be efficiently adapted to various scenarios through simple fine-tuning.
The details of the datasets and models will be presented in the subsequent sections.

\subsection{Formula Recognition Dataset}
Existing OCR datasets mostly focus on formulas with simple symbols and structures. 
They mainly target pure formulas and fail to fully consider the complex layout structures with interleaved text and formulas that are widely present in actual documents.
Thus, the specialized models trained on these public datasets show obvious bottlenecks when dealing with multi-level nested structures, complex subscripts and superscripts, special operators, and interdisciplinary symbol systems. 
They struggle to handle the rich structures and diverse expressions of advanced formulas in fields such as physics, chemistry, and biology. 

To address the above challenges, we independently design efficient data crawling and processing tools. 
We systematically collect, clean, and organize a large number of complex formula samples from academic resources across multiple disciplines, and construct a large-scale and challenging dataset, CSFormula, which encompasses diverse scientific fields, various structural types, and complex layouts.

\noindent \textbf{Dataset Overview.} The proposed CSFormula dataset comprises a large collection of image-LaTeX pairs, encompassing formulas with diverse structures and spanning multiple disciplines, including mathematics, physics, and chemistry. Notably, the dataset features complex layout structures—such as multi-line and page-level formulas—thereby more accurately reflecting the variety and challenges of formulas encountered in scientific literature. Specifically, the CSFormula dataset is organized into three distinct categories:
\begin{itemize}
    \item \textbf{Line-level:} Contains individual printed formulas, including both single-line and multi-line expressions. This subset is designed to evaluate a model's ability to parse and interpret formulas of varying lengths and complexities.
    \item \textbf{Paragraph-level:} Focuses on scenarios where formulas are embedded within natural language paragraphs, enabling assessment of a model's capability to segment and recognize formulas in mixed-content documents.
    \item \textbf{Page-level:} Consists of full-page scientific document images that include formulas, text, tables, and other elements. This subset tests the model's robustness in handling complex layouts and page-level structures.
\end{itemize}

\noindent \textbf{Data Collection Process.} We began by crawling 5,827,854 web pages from the StackExchange website, noting that some pages might be duplicated during the collection process. To identify relevant content, we employed regular expressions to identify pages containing LaTeX formulas and subsequently counted, prior to deduplication, the number of line-level, paragraph-level, and page-level formulas (see Table~\ref{tab:dedup}). We then performed a thorough deduplication process to ensure the uniqueness of the collected data. Finally, we constructed HTML templates and utilized MathJax to render the identified LaTeX formulas into images, producing the final set of image-LaTeX pairs. This unified pipeline was applied consistently across line-level, paragraph-level, and page-level formulas, drawing on data from official community sources and STEM-related channels. After deduplication, the dataset contains 741,016 line-level, 135,575 paragraph-level, and 131,876 page-level samples for training, and 1,000 samples at each level for testing.

\subsection{Vision-language Model}
Most formula recognition methods employ dedicated models that rely on structural customization and extensive prior knowledge, which limits the generalization ability of these models across different formula recognition scenarios.
Unlike these models that depend on structural customization, we directly utilize general VLMs such as Qwen2.5-VL~\cite{Qwen2.5-VL} and perform domain adaptation on the complex formula recognition dataset through SFT. 
Benefiting from the strong generalization ability of large models, the fine-tuned model achieves SOTA recognition performance in various complex scenarios without the need for sophisticated structural design and prior knowledge.
We detail the structure and loss below.

\noindent \paragraph{Structure}:
Large-scale multimodal models, such as Qwen2.5-VL~\cite{Qwen2.5-VL}, aim to effectively integrate the capabilities of pre-trained Large Language Models (LLMs) and visual models. 
Specifically, the vision encoder takes a literature image $\mathbf{v} \in \mathbb{R}^{H \times W \times 3}$ as input, where $H$ and $W$ represent the height and the width, respectively. 
Before being fed into the ViT, the height and width of the input image are resized to $H'$ and $W'$ (multiples of 28). 
Then, the ViT processes images by splitting them into patches with a stride of 14, generating a set of image features. 
Finally, the visual encoder outputs a visual token sequence $\mathcal{Z} = [z_1, z_2, \cdots, z_{P_0}] \in \mathbb{R}^{P_0 \times C}$, with $P_0$ denoting the length of the visual token sequence. 
The vision encoder incorporates 2D-RoPE and window attention to support native input resolutions while accelerating the computation.

Given vision tokens, instead of directly using these tokens, Qwen2.5-VL first groups spatially adjacent sets of four tokens. 
These grouped features are then concatenated and passed through a two-layer multi-layer perceptron (MLP) to map them to a vision token sequence $\mathcal{V} \in \mathbb{R}^{P \times D}$, where $D$ is the hidden size of the LLM.

Additionally, the instruction text is projected into instruction text tokens $\mathcal{T} = [t_1, t_2, \cdots, t_N] \in \mathbb{R}^{N \times D}$, where $N$ is the length of the instruction text token sequence. 
The pre-trained large language model consists of stacked multi-head self-attention (MSA) and feed-forward networks (FFN), with each block typically incorporating RMSNorm and residual connections:
\begin{equation}
    \mathbf{x}_0 = [v_1, v_2, \cdots, v_P, \cdots, t_1, t_2, \cdots, t_N],
\end{equation}
\begin{equation}
    \mathbf{x}_{\ell}^{\prime} = \mathrm{MSA}(\mathrm{RMSNorm}(\mathbf{x}_{\ell-1})) + \mathbf{x}_{\ell-1}, \quad \ell \in \{1, \ldots, L\},
    \label{eq:msa}
\end{equation}
\begin{equation}
    \mathbf{x}_{\ell} = \mathrm{FFN}(\mathrm{RMSNorm}(\mathbf{x}^{\prime}_{\ell})) + \mathbf{x}^{\prime}_{\ell}, \quad \ell \in \{1, \ldots, L\},
    \label{eq:ffn}
\end{equation}
where $L$ is the number of layers in the LLM.
Unlike typical methods that normalize coordinates, Qwen2.5-VL directly uses the actual dimensions of the input image (absolute coordinates) to represent bounding boxes, points, and other spatial features. 
This allows the model to learn scale information inherently, improving its ability to process images across different resolutions.

\noindent \paragraph{Loss}:
The outputs are optimized using a generative loss in an autoregressive manner:
\begin{equation}
    \mathcal{L}_{\text{main}}(\theta) = -\sum_{i=1}^K \log p\left(y_i \mid \mathcal{V}, \mathcal{T}, \mathcal{Y}_{<i}; \theta \right),
    \label{eq:gen}
\end{equation}
where $\mathcal{Y}_{<i}$ indicates output LaTeX formula tokens $[y_{1},y_{2},\cdots,y_{i-1}]$ ($i\geq2$) and no output LaTeX formula tokens when $i=1$. 
$\theta$ indicates the trainable parameters.

\section{Experiments}
\label{sec:experiments}

\subsection{Experimental Settings}
To ensure a comprehensive and rigorous assessment, we conduct experiments on a diverse set of benchmark datasets that cover different real-world scenarios, academic domains, and structural complexities. These datasets encompass a wide range of formula styles and layouts, providing a robust foundation for evaluating both the effectiveness and the generalization ability of our approach. Furthermore, we employ multiple evaluation metrics—such as Edit Distance (ED)~\cite{lcvenshtcin1966binary} and Character Detection Matching (CDM)~\cite{wang2025image}—to thoroughly measure performance across various aspects of formula understanding.

\begin{table*}[!ht]
    \centering
    \resizebox{\textwidth}{!}{
    \begin{tabular}{l l c cccc ccccc}
        \toprule
        \textbf{Model Type} & \textbf{Methods} 
        & \multicolumn{1}{c}{\textbf{Im2LaTeX-160k}} 
        & \multicolumn{5}{c}{\textbf{UniMER}} 
        & \multicolumn{4}{c}{\textbf{CSFormula }} \\
        \cmidrule(lr){3-3} \cmidrule(lr){4-8} \cmidrule(lr){9-12}
        & & SPE$\downarrow$ 
        & SPE$\downarrow$ & CPE$\downarrow$ & SCE$\downarrow$ & HWE$\downarrow$ & Avg.$\downarrow$
        & Line$\downarrow$ & Paragraph$\downarrow$ & Page$\downarrow$ & Avg.$\downarrow$  \\
        \midrule
        Task-Specific (SOTA) & UniMERNet & \textbf{0.240} & \textbf{0.060} & \textbf{0.056} & \underline{0.224} & \underline{0.072} & \underline{0.103} & 0.489 & 0.645 & 0.903 & 0.679 \\
        Open VLM (SOTA) & Qwen2.5-VL & 0.310 & 0.185 & 0.460 & 0.364 & 0.206 & 0.303 & 0.416 & \underline{0.315} & 0.684 & 0.472  \\
        Close VLM (SOTA) & GPT-4o & 0.434 & 0.497 & 0.528 & 0.644 & 0.512 & 0.545 & 0.338 & 0.357 & 0.511 & 0.402\\
        Close VLM (SOTA) & Gemini-2.5-flash & 0.424 & 0.481 & 0.571 & 0.601 & 0.472 & 0.531 & \underline{0.312} & 0.378 & \underline{0.494} & \underline{0.394} \\
        Professional Tool (SOTA) & Mathpix & 0.449 & 0.467 & 0.474 & 0.589 & 0.535 & 0.516 & 0.407 & 0.446 & 0.518 & 0.457 \\
        \hline
        & DocTron-Formula & \underline{0.245} & \underline{0.081} & \underline{0.084} & \textbf{0.182} & \textbf{0.046} & \textbf{0.098}  & \textbf{0.121} & \textbf{0.121} & \textbf{0.251} & \textbf{0.164} \\
        \bottomrule
    \end{tabular}
    }
    \vspace{-2mm}
    \caption{
    Comparison of formula recognition performance on three benchmarks. 
    All metrics are ED, lower is better. 
    The best results for each column are highlighted in \textbf{bold}, and the second best are \underline{underlined}.
    }
    \vspace{+1mm}
    \label{tab:ed_model_comparison}
\end{table*}

\begin{table*}[!ht]
    \centering
    \resizebox{\textwidth}{!}{
    \begin{tabular}{l l c ccccc ccccc}
        \toprule
        \textbf{Model Type} & \textbf{Method}
        & \multicolumn{1}{c}{\textbf{Im2LaTeX-160k}}
        & \multicolumn{5}{c}{\textbf{UniMER}} 
        & \multicolumn{4}{c}{\textbf{CSFormula}} \\
        \cmidrule(lr){3-3} \cmidrule(lr){4-8} \cmidrule(lr){9-12}
        & & SPE$\uparrow$
        & SPE$\uparrow$ & CPE$\uparrow$ & SCE$\uparrow$ & HWE$\uparrow$ & Avg.$\uparrow$ 
        & Line$\uparrow$ & Paragraph$\uparrow$ & Page$\uparrow$ & Avg.$\uparrow$
        \\
        \midrule
        Task-Specific (SOTA) & UniMERNet  & \textbf{0.991} & \textbf{0.994}  & \textbf{0.970}  & 0.946  & \textbf{0.953}  & \textbf{0.965}  & 0.919  & 0.644  & 0.009  & 0.524  \\
        Open VLM (SOTA) & Qwen2.5-VL  & 0.971  & 0.952  & 0.818  & \underline{0.947}  & 0.927  & 0.911 & {0.924}  & \underline{0.746}  & 0.197  & 0.622 \\
        Close VLM (SOTA) & GPT4o  & 0.929  & 0.786  & 0.641  & 0.866  & 0.842  & 0.783  & 0.879  & 0.569  & 0.161  & 0.536 \\
        Close VLM (SOTA) & Gemini-2.5-flash  & {0.973} & {0.965}  & 0.739  & 0.929  & 0.898  & 0.882  & 0.880  & {0.725}  & \underline{0.592}  & {0.732}  \\
        Professional Tool (SOTA) & Mathpix  & 0.969  & 0.973  & \underline{0.967}  & 0.932  & 0.924  & 0.949  & \underline{0.926}  & 0.696  & {0.579}  & \underline{0.733} \\
        \hline
        & DocTron-Formula  & \underline{0.985} & \underline{0.979}  & 0.962  & \textbf{0.958}  & \underline{0.947}  & \underline{0.961} & \textbf{0.950}  & \textbf{0.897}  & \textbf{0.774}  & \textbf{0.873}  \\
        \bottomrule
    \end{tabular}
    }
    \vspace{-2mm}
    \caption{
        Comparison of formula recognition performance on three benchmarks. 
        All metrics are CDM, higher is better. 
        The best results for each column are highlighted in \textbf{bold}, and the second best are \underline{underlined}.
    }
    \vspace{-3mm}
    \label{tab:cdm_model_comparison}
\end{table*}

\subsubsection{Datasets}
We evaluate our method on the following datasets:

\textbf{UniMER Dataset.} The UniMER~\cite{wang2024unimernet} test set contains 23,757 samples, divided into four types of expressions: 6,762 Simple Printed Expressions (SPE), 5,921 Complex Printed Expressions (CPE), 4,742 Screen-Captured Expressions (SCE), and 6,332 Handwritten Expressions (HWE). 

\textbf{Im2LaTeX-160K Dataset.} The Im2LaTeX-160K\footnote{\url{https://huggingface.co/datasets/yuntian-deng/im2latex-100k/tree/main}} dataset, a widely used benchmark for mathematical formula recognition, contains printed formulas from scientific literature paired with LaTeX annotations and includes a test set of 20,430 samples.

\textbf{CSFormula Dataset.} The CSFormula dataset contains a large collection of LaTeX-image pairs with diverse structures from mathematics, physics, chemistry, and other disciplines. It features complex layouts at the line, paragraph, and page levels, closely reflecting the challenges found in scientific literature. The test set includes 1,000 samples for each level, providing a comprehensive benchmark for real-world model evaluation.

\subsubsection{Metrics}
To comprehensively evaluate our method, we adopt both traditional and novel metrics:

\textbf{Edit Distance.} This metric calculates the minimum number of edit operations (insertions, deletions, substitutions) needed to convert the predicted LaTeX string into the ground truth, reflecting character-level accuracy. While widely used for fair comparison with previous works, it may introduce bias when structurally different but mathematically equivalent formulas are present.

\textbf{Character Detection Matching.} CDM~\cite{wang2025image} is a visual-space metric that renders both predicted and ground-truth formulas as images and computes character-level overlap. By evaluating visual similarity, CDM effectively mitigates the biases introduced by different but equivalent LaTeX representations, resulting in a more consistent and human-aligned assessment of formula recognition performance. Its effectiveness has been demonstrated in~\cite{wang2025image}.


\subsection{Performance Comparisons}
We conduct a comprehensive comparison of formula recognition performance across a diverse set of representative models, each with distinct design philosophies and technical characteristics. The evaluated methods include our DocTron-Formula, which adopts a large-scale model architecture and is further fine-tuned specifically for mathematical formula recognition. The task-specific model UniMERNet~\cite{wang2024unimernet} is tailored for formula understanding through domain-specific training on datasets encompassing diverse styles, including printed, screenshot, and handwritten formulas. Open-source and closed-source general VLMs (e.g., Qwen2.5-VL~\cite{Qwen2.5-VL}, GPT-4o~\cite{openai2024gpt4o}, and Gemini-2.5-flash~\cite{comanici2025gemini}) are designed for broad cross-domain visual-linguistic tasks but are not exclusively optimized for mathematical content. Additionally, Mathpix\footnote{\url{https://mathpix.com/equation-to-latex}} serves as a professional OCR tool widely used in academic and industrial scenarios, recognized for its strong practical performance in formula recognition.

\begin{figure*}[!htbp]
\centering
\includegraphics[width=\textwidth]{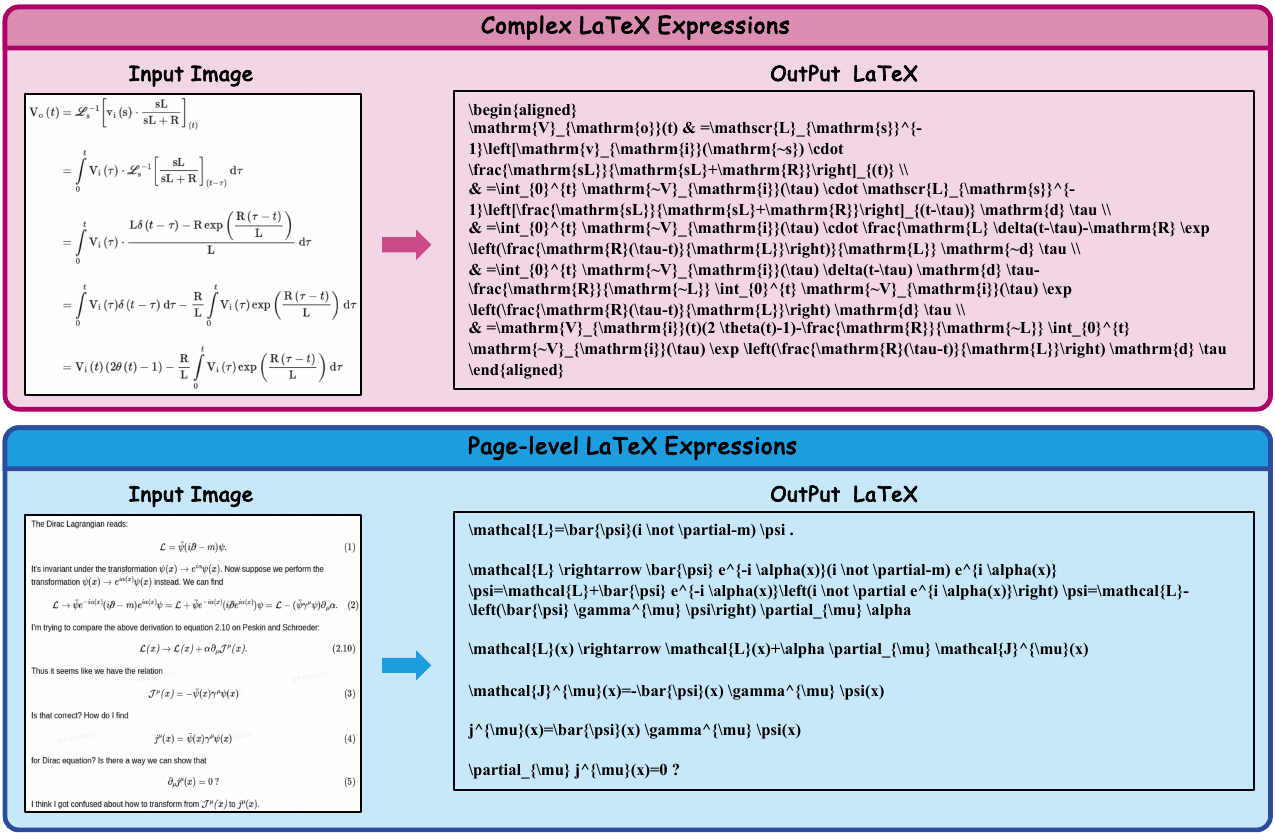}
\vspace{-8mm}
\caption{Visualization of DocTron-Formula’s recognition results. The top shows accurate recognition of complex, deeply nested LaTeX expressions, while the bottom demonstrates robust parsing of page-level layouts.}
\vspace{-2mm}
\label{fig:qualitative}
\end{figure*}

The results in Table~\ref{tab:ed_model_comparison} demonstrate the remarkable capability of DocTron-Formula to handle structural complexity and generalization challenges as measured by the ED metric. On the public benchmarks, DocTron-Formula exhibits highly competitive performance: for Im2LaTeX-160k, it achieves an ED of 0.245, closely matching the task-specific SOTA UniMERNet (0.240) and outperforming all other baselines. On UniMER, it achieves the lowest average ED (0.098), surpassing UniMERNet (0.103) and excelling particularly in the most challenging subsets, such as Screen-Captured Expressions (SCE, 0.182) and Handwritten Expressions (HWE, 0.046). These results underscore DocTron-Formula’s robustness in handling both synthetic and real-world formula variations. 
On the comprehensive CSFormula dataset, DocTron-Formula demonstrates a substantial advantage in generalization, achieving the lowest average ED (0.164) and outperforming the second-best method, Gemini-2.5-flash (0.394), by a large margin. DocTron-Formula consistently outperforms UniMERNet at all levels, indicating that UniMERNet’s performance declines notably in more realistic and structurally complex scenarios.
Furthermore, general-purpose vision-language models (Qwen2.5-VL, GPT-4o, Gemini-2.5-flash) exhibit better generalization than UniMERNet on CSFormula, but still lag behind DocTron-Formula.

The results in Table~\ref{tab:cdm_model_comparison} highlight DocTron-Formula’s superiority in terms of visual similarity as measured by the CDM metric.
On public benchmarks, DocTron-Formula achieves top-tier visual recognition: for Im2LaTeX-160k, it reaches an accuracy of 0.985, closely matching UniMERNet (0.991) and outperforming all other baselines. On UniMER, DocTron-Formula attains an average score of 0.961, just 0.004 below UniMERNet (0.965), and achieves the highest score on SCE (0.958), reflecting its ability to produce visually faithful results that align well with human perception.
On the challenging CSFormula dataset, DocTron-Formula achieves the highest average CDM score (0.873), outperforming the second-best method, Mathpix (0.733), by a notable margin. Its superior performance extends to the most difficult scenarios, with paragraph-level and page-level scores of 0.897 and 0.774, respectively, while UniMERNet lags behind with scores of 0.644 and 0.009. This demonstrates DocTron-Formula’s capability to maintain visual accuracy and consistency even in highly diverse and realistic document layouts.
While general vision-language models (Qwen2.5-VL, GPT4o, Gemini-2.5-flash) do not reach the level of DocTron-Formula, they exhibit better adaptability than UniMERNet in diverse, real-world settings.

Overall, these results highlight that DocTron-Formula not only matches the task-specific SOTA model on standard benchmarks, but also sets a new state-of-the-art in complex and generalized formula recognition tasks, demonstrating superior robustness and generalization across diverse real-world scenarios.

\subsection{Qualitative Results}

To further demonstrate the effectiveness and robustness of DocTron-Formula, we present qualitative results across a variety of representative scenarios. These visualizations provide intuitive insights into the model’s ability to handle diverse structural layouts, adapt to different formula styles, recover complex mathematical expressions, and generalize across scientific domains.

\vspace{-2mm}
\noindent \paragraph{Structural Complexity and Cross-domain Adaptability Analysis.}
To illustrate DocTron-Formula’s ability to handle both structural complexity and cross-domain generalization, we visualize recognition results on complex, multi-line LaTeX expressions from advanced scientific domains. As illustrated at the \textit{top} of Fig.~\ref{fig:qualitative}, the visualizations confirm that DocTron-Formula can faithfully recover both symbol content and hierarchical structure in challenging scientific expressions. This demonstrates the model’s superior structural understanding and strong adaptability to a wide range of scientific formula conventions.

\begin{table*}[!ht]
    \centering
    \resizebox{0.8\textwidth}{!}{
    \begin{tabular}{l ccc ccc}
        \toprule
        \textbf{Exp} 
        & \multicolumn{3}{c}{\textbf{Training Data}} 
        & \multicolumn{3}{c}{\textbf{Testing Data}} \\
        \cmidrule(lr){2-4} \cmidrule(lr){5-7}
        & Line-level & Paragraph-level & Page-level 
        & Line-level $\downarrow$ & Paragraph-level $\downarrow$ & Page-level $\downarrow$ \\
        \midrule
        1 & \Checkmark & {\color[gray]{0.8} \ding{55}} & {\color[gray]{0.8} \ding{55}} & 0.123 & --    & --    \\
        2 & {\color[gray]{0.8} \ding{55}} & \Checkmark & {\color[gray]{0.8} \ding{55}} & --    & 0.137 & --    \\
        3 & {\color[gray]{0.8} \ding{55}} & {\color[gray]{0.8} \ding{55}} & \Checkmark & --    & --    & 0.278 \\
        4 & \Checkmark & \Checkmark & {\color[gray]{0.8} \ding{55}} & 0.122 & 0.126 & --    \\
        5 & \Checkmark & \Checkmark & \Checkmark & \textbf{0.121} & \textbf{0.123} & \textbf{0.272} \\
        \bottomrule
    \end{tabular}
    }
    \vspace{-1mm}
    \caption{Ablation study on the impact of different training data levels for DocTron-Formula.}
    \label{tab:ablation_multilevel}
\end{table*}

\begin{table*}[!ht]
    \centering
    \resizebox{0.8\textwidth}{!}{
    \begin{tabular}{l c ccccc ccccc}
        \toprule
        \textbf{Model} & \textbf{Size}
        & \multicolumn{1}{c}{\textbf{Im2LaTeX-160k}}
        & \multicolumn{5}{c}{\textbf{UniMER-Test}} 
        & \multicolumn{4}{c}{\textbf{CSFormula}}
        \\
        \cmidrule(lr){3-3} \cmidrule(lr){4-8} \cmidrule(lr){9-12}
        & & SPE$\downarrow$
        & SPE$\downarrow$ & CPE$\downarrow$ & SCE$\downarrow$ & HWE$\downarrow$ & Avg.$\downarrow$
        & Line$\downarrow$ & Paragraph$\downarrow$ & Page$\downarrow$ & Avg.$\downarrow$
         \\
        \midrule
        Qwen2.5-VL & 3B & 0.322 & 0.190 & \textbf{0.235} & 0.436 & 0.226 & \textbf{0.271}  & 0.475 & 0.539 & 0.843 & 0.619 \\
        Qwen2.5-VL & 7B & \textbf{0.310} & \textbf{0.185} & 0.460 & \textbf{0.364} & \textbf{0.206} & {0.303}  & \textbf{0.416} & \textbf{0.315} & \textbf{0.684} & \textbf{0.472} \\
        \midrule
        DocTron-Formula & 3B & 0.253 & 0.095 & 0.099 & \textbf{0.177} & 0.055 & 0.106  & 0.122 & 0.131 & 0.297 & 0.183 \\
        DocTron-Formula & 7B & \textbf{0.245} & \textbf{0.081} & \textbf{0.084} & 0.182 & \textbf{0.046} & \textbf{0.098} & \textbf{0.121} & \textbf{0.121} & \textbf{0.251} & \textbf{0.164} \\
        \bottomrule
    \end{tabular}
    }
    \vspace{-1mm}
    \caption{Comparison of different model sizes and methods on three benchmarks. All metrics are ED, lower is better.}
    \vspace{-1mm}
    \label{tab:model_size_comparison}
\end{table*}

\vspace{-2mm}
\noindent \paragraph{Visualization of Page-Level.}
We qualitatively assess the structural adaptability of DocTron-Formula by visualizing its recognition results on page-level academic documents containing multiple complex and multi-line formulas. As illustrated at the \textit{bottom} of Fig.~\ref{fig:qualitative}, the results show that DocTron-Formula consistently achieves accurate structural parsing and symbol alignment, even in challenging scenarios with dense layouts and diverse scientific expressions. The model demonstrates strong capability in maintaining structural integrity and correctly associating symbols across different formulas on the same page.

\vspace{-1mm}
\subsection{Ablation Studies}
To further clarify the factors underlying DocTron-Formula's performance, we conduct two complementary ablation studies. Specifically, we first explore how the combination of line-level, paragraph-level, and page-level data during training contributes to improved generalization across diverse formula structures. Then, we examine the effect of model size by comparing recognition accuracy at different parameter scales. 

\vspace{-1mm}
\noindent \paragraph{Impact of Hierarchical Training Data on Model Performance.}
Table~\ref{tab:ablation_multilevel} presents an ablation study on how training data at different structural levels affects DocTron-Formula’s performance. We varied the inclusion of line-, paragraph-, and page-level data during training and evaluated on corresponding test sets. The results show that models trained and tested at the same structural level (Exp 1--3) have limited performance. In contrast, training with multiple structural levels (Exp 4) improves results on both line- and paragraph-level test sets (0.122 and 0.126). The best performance is achieved when all three levels are included (Exp 5), resulting in the lowest errors across all test sets (0.121, 0.123, 0.272). These findings demonstrate that joint training with diverse structural data enhances the model’s generalization and robustness.

\noindent \paragraph{Impact of Model Size on Formula Recognition.}
Table~\ref{tab:model_size_comparison} shows the impact of model size on formula recognition performance. For both Qwen2.5-VL and DocTron-Formula, increasing model size from 3B to 7B leads to consistent improvements across most evaluated benchmarks and metrics. For example, Qwen2.5-VL improves from 0.322 to 0.310 SPE on Im2LaTeX-160k, and from 0.619 to 0.472 on CSFormula average. Similarly, DocTron-Formula improves from 0.253 to 0.245 SPE on Im2LaTeX-160k, from 0.106 to 0.098 on UniMER average, and from 0.183 to 0.164 on CSFormula average.
Notably, even at the 3B scale, DocTron-Formula consistently outperforms Qwen2.5-VL (7B) on all benchmarks (e.g., on Im2LaTeX-160k, DocTron-Formula achieves 0.253 SPE vs. 0.310 for Qwen2.5-VL; on UniMER average, 0.106 vs. 0.303; and on CSFormula average, 0.183 vs. 0.472), underscoring the strength of our approach. Overall, larger model sizes result in better formula recognition, and DocTron-Formula demonstrates clear advantages over Qwen2.5-VL at both scales.

\vspace{-2mm}
\section{Conclusion}
\label{sec:conclusion}
In this work, we explored generalized formula recognition in complex and structured scenarios, leveraging general multimodal models and a structurally diverse dataset to achieve state-of-the-art performance across challenging layouts and styles. However, our current focus remains on mathematical formulas, while broader scientific document understanding—such as the recognition of tables, figures, and the modeling of complex reading orders—remains an open challenge. Future work will extend our approach to holistic document analysis, aiming to integrate formula recognition with comprehensive extraction and interpretation of various document elements, and to enhance semantic understanding within full scientific texts.
{
    \small
    \bibliographystyle{ieeenat_fullname}
    \bibliography{main}
}


\end{document}